\documentclass{article}

\usepackage{fancyhdr}

\usepackage{arxiv}

\usepackage[utf8]{inputenc} 
\usepackage[T1]{fontenc}    
\usepackage{hyperref}       
\usepackage{url}            
\usepackage{booktabs}       
\usepackage{amsfonts}       
\usepackage{nicefrac}       
\usepackage{microtype}      
\usepackage{lipsum}		
\usepackage{graphicx}
\usepackage{doi}

\usepackage[utf8]{inputenc} 
\usepackage[T1]{fontenc}    
\usepackage{hyperref}       
\usepackage{url}            
\usepackage{booktabs}       
\usepackage{amsfonts}       
\usepackage{nicefrac}       
\usepackage{microtype}      
\usepackage{lipsum}
\usepackage{fancyhdr}       
\usepackage{graphicx}       
\graphicspath{{media/}}     

\usepackage{longtable}
\usepackage{float}
\usepackage{adjustbox}
\usepackage{multirow}
\usepackage{tabularx}
\usepackage{rotating} 
\usepackage{makecell}  
\usepackage{caption}
\usepackage{xcolor} 

\usepackage{pifont}

\NewDocumentCommand\xtnote{sm}{\def\tss{\textsuperscript{#2}}\IfBooleanTF{#1}{\tss}{\rlap{\tss}}}

\newcolumntype{M}[1]{>{\centering\arraybackslash}m{#1}}

\title{Deep Learning Techniques for Hyperspectral Image Analysis in Agriculture: A Review}

\author{
  Mohamed Fadhlallah Guerri \textsuperscript{1,2}, Cosimo Distante \textsuperscript{1,2}, Paolo Spagnolo \textsuperscript{2},  Fares Bougourzi \textsuperscript{3}, and Abdelmalik Taleb-Ahmed \textsuperscript{4}\\
  \textsuperscript{1}Department of Innovation Engineering, University of Salento, 73100 Lecce, Italy\\
  \textsuperscript{2}National Research Council of Italy \\, Institute of Applied Sciences and Intelligent Systems, via Monteroni snc 73100 Lecce, Italy\\
  \textsuperscript{3}University Paris-Est Creteil, Laboratoire LISSI \\, 122 Rue Paul Armangot, Vitry sur Seine, 94400, Paris, France\\
  \textsuperscript{4}Institute d'Electronique de Microelectronique et  \\ de Nanotechnologie (IEMN), UMR 8520, Universite Polytechnique Hauts de France,  \\ Universite de Lille, CNRS, 59313, Valenciennes, France\\
  \texttt{\{mohamedfadhlallah.guerri, cosimo.distante\}@unisalento.it}, \\  \texttt{paolo.spagnolo@cnr.it}, \texttt{faresbougourzi@gmail.com}, \texttt{Abdelmalik.Taleb-Ahmed@uphf.fr}
}

\begin{document}

\maketitle

\begin{abstract}
In the recent years, hyperspectral imaging (HSI) has gained considerably popularity among computer vision researchers for its potential in solving remote sensing problems, especially in agriculture field. However, HSI classification is a complex task due to the high redundancy of spectral bands, limited training samples, and non-linear relationship between spatial position and spectral bands. Fortunately, deep learning techniques have shown promising results in HSI analysis. This literature review explores recent applications of deep learning approaches such as Autoencoders, Convolutional Neural Networks (1D, 2D, and 3D), Recurrent Neural Networks, Deep Belief Networks, and Generative Adversarial Networks in agriculture. The performance of these approaches has been evaluated and discussed on well-known land cover datasets including Indian Pines, Salinas Valley, and Pavia University.

\end{abstract}



\section{Introduction}\label{Introduction}

In the last 20 years, there has been an increasing need to assess the quality and safeguarding of horticultural and agricultural produce. With the advent of sophisticated agricultural technologies, this has become an indispensable aid for farmers in managing crop health and resource utilization \cite{sethy2021hyperspectral}. Traditional approaches for obtaining crop classification results through field measurement, investigation, and statistics are time-consuming, labor-intensive, and expensive \cite{ravikanth2017extraction}. Therefore, a non-destructive, non-polluting, and quick technology such as Hyperspectral Imaging (HSI) has emerged as a potential solution. HSI has the potential to capture multiple images across different wavelengths, enabling precise monitoring of spatial and temporal variations in farmland. This capability facilitates rapid and accurate predictions of crop growth \cite{ge2016temporal}. HSI finds diverse applications in agriculture, ranging from crop management \cite{zhang2016crop}, forecasting crop yield \cite{li2020above}, and detecting crop diseases \cite{mahlein2013development} to monitoring land usage \cite{selige2006high}, water resources \cite{gursoy2015determining}, and soil conditions \cite{weber2008new}. Deep learning methods have shown promising results in many agricultural applications, enabling farmers to make crucial decisions when needed. They offer a number of benefits over traditional Machine Learning (ML) methods, including the ability to automatically extract highly relevant characteristics. Crop classification tasks have seen significant growth in the use of deep learning algorithms in recent years, with several significant efforts to enhance this problem utilizing current deep learning algorithms \cite{bouguettaya2022deep}.


HSI has revolutionized the way farmers approach agriculture by enabling them to make quick and informed decisions about their crops. The integration of deep learning algorithms with HSI has enhanced the accuracy and efficiency of crop classification and other agricultural applications, leading to a more sustainable and profitable agricultural industry. The recent advancements in Artificial Intelligence (AI) have led to the integration of AI techniques with various applications in research and the business world. This integration has opened up new possibilities for the development of smart systems in horticultural, agricultural, and food domains, especially since The rise of ML, a sub-branch of AI that deals with algorithms that learn to recognize patterns in data to make decisions \cite{nturambirwe2020machine}. The analysis and interpretation of enormous volumes of data produced by Hyperspectral Imaging (HSI) systems present numerous difficulties and continue to be a bottleneck in many horticultural and agricultural applications. However, AI approaches, particularly Deep Learning (DL), can use the depth of the spectral and spatial information to identify correlations with quality parameters when applied to HSI data \cite{signoroni2019deep}. DL has become increasingly important due to its superior efficiency and quality compared to conventional machine learning models \cite{vantaggiato2021covid, jaiswal2021critical, bougourzi2023cnr, bougourzi2022deep}. Combining hyperspectral data with cutting-edge AI approaches, especially DL, offers a wide range of possibilities for fresh product quality management.

Several studies have highlighted the potential of HSI in the agricultural section, the use of ML algorithms for data analysis and interpretation has also been extensively explored. Table \ref{Tab:surveysdiscussion} represents some of previous review papers contribution of HSI in the field of agriculture.

The primary contribution of this paper is to examine the application of HSI technology and to close gaps in the study of HSI systems. This review paper highlights the advantages of using HSI over RGB cameras, such as the substantial quantity of data acquired in a single image, which is not visible to the human eye. Additionally, the paper discusses the use of Deep Learning (DL), which can offer a superior efficiency and quality for product quality management. By combining HSI data with DL techniques, this paper provides insights into new possibilities for the development of smart systems in horticultural, agricultural, and food domains, and contributes to the advancement of the use of AI in the industry.
This paper's main contributions can be summed up as follows:

\begin{itemize}
    \item Firstly, this paper provides a comprehensive discussion of the general concepts and essential information related to HSI technology and imaging methods. This discussion is intended to enhance the understanding of the technology and its potential applications for young researchers.
    
    \item Secondly, the paper analyzes various publicly available HSI agricultural datasets, highlighting their unique features, and how they have been utilized in agriculture.
    
    \item Thirdly, the study reviews different techniques and approaches of deep learning used for HSI, providing an in-depth analysis of their strengths and limitations.
    
    \item Lastly, the paper presents and discusses the main applications in the field of agriculture that benefit from HSI technology. This section provides insights into how HSI can be used to improve crop yield, quality, and safety.
\end{itemize}

Section \ref{HSI} of this article provides basic information on hyperspectral imaging and tools and explication. Section \ref{acqmodes} discusses four common methods of acquiring hyperspectral images, while section \ref{datasets} presents some publicly available datasets. Section \ref{techniques} analyzes various approaches and techniques of deep learning utilized in hyperspectral imaging technology. In Section \ref{apps}, different approaches to the application of hyperspectral imaging technology in agriculture are discussed. Finally, the article concludes with a discussion of limitations and potential areas for future research.

\begin{table*}
\centering
\begin{minipage}{\textwidth}
\caption{Summary of Surveys and reviews related to HSI in agriculture with comparison}\label{Tab:surveysdiscussion}
\begin{adjustbox}{width=\textwidth}

\begin{tabular}{ l l l l p{0.5\textwidth}}
 \hline

Reference & Imaging methods & Applications of HSI & DL techniques & Description \\ [0.5cm] \hline

\cite{lu2020recent} & \checkmark & \checkmark & \ding{55} & Provides an in-depth discussion of imaging platforms, sensors, and analytical methods used in agricultural research. The effectiveness of hyperspectral imaging for various applications. \\

\cite{zhang2020review} & \ding{55} & \checkmark & \ding{55} & Elaborates on the significant benefits of hyperspectral technologies in detecting plant diseases. Outlines the steps involved in analyzing hyperspectral disease data, including the various articles, algorithms, and methods used for disease detection. \\

\cite{wang2021review} & \ding{55} & \checkmark & \checkmark & Overview of the numerous applications of hyperspectral imaging (HSI) in the agricultural sector, such as predicting ripeness and components, classifying different themes, and detecting plant diseases. Additionally, it discusses the recent advancements and difficulties faced by deep learning models and feature networks in this field. 
\\

\cite{terentev2022current} & \ding{55} & \checkmark & \checkmark & Discusses recent developments in hyperspectral remote sensing-based early plant disease detection and identifying existing shortcomings in experimentation methodology of plant disease detection.  \\

Ours & \checkmark & \checkmark & \checkmark & This article delves into the latest uses of deep learning techniques within the agriculture industry. The effectiveness of these methods has been examined and analyzed using prominent land cover datasets. Also covers the imaging methods and summarizes various HSI applications in agriculture. \\
 \hline

\end{tabular}
\end{adjustbox}
\end{minipage}
\end{table*}

\pagebreak

\section{Hyperspectral Imaging}\label{HSI}

Hyperspectral imaging (HSI or HI), also known as chemical and spectroscopic imaging, is a novel technique has been developed that merges conventional imaging with spectroscopy, allowing for the simultaneous acquisition of both spatial and spectral data from an image. Despite being originally designed for remote sensing, HSI technology's advantages over traditional machine vision are now evident in diverse fields, including agriculture. Optical sensing and imaging techniques have advanced to the point where HSI is now a valuable method for technical inspection and consistency measurement of fruits and vegetables.

\subsection{Imaging sensor}

The HSI system is composed of four crucial elements, namely the illumination source, primary lens, region of interest (ROI) detector, and spectroscopic imager. Choosing an appropriate illumination source is essential for optimal performance and reliability. Choosing an appropriate objective lens relies on the capability to focus gathered light originating from a limited region onto the detector unit, resulting in the formation of pixels in the output image. Achieving appropriate spatial resolution is critical in HSI systems and is determined by the optical input slot volume with respect to wavelength and the detector component size of primary lenses \cite{edelman2012hyperspectral}. The spectrograph receives the light from the objective lenses and disperses it into separate wavelengths. This is accomplished through the use of imaging spectrographs that rely on diffraction gratings, which consist of evenly spaced grooves and play a significant role in scattering wavelengths \cite{woodgate1998space}. In the end, the dispersed light is captured by a detector that transforms photons into electrical signals. These signals are analyzed by a computer to determine the intensity rates of different wavelengths. Two main types of solid-state region detectors, the charge-coupled device (CCD) and the complementary metal-oxide semiconductor (CMOS), are utilized as image sensors. \cite{el2005cmos}.

\subsection{Hyperspectral imaging data acquisition}
HSI aims to capture the spectral range of each pixel in an image of a scene to facilitate substance identification, target detection, and processing \cite{chang2003hyperspectral}. High spectral resolution and small band images are typically produced using a combination of hyperspectral images, spectroscopic methods, 2-dimensional geometric space, and 1-dimensional spectral detail detection. Through extensive research and development, HSI has found many useful applications in the quality assessment of precision agriculture. HSI combines spectroscopic and imaging techniques into a single device that can obtain the spatial map of spectral variation \cite{sethy2021hyperspectral}. 

\begin{center}
    \begin{figure}[h!]
        \includegraphics[width=\linewidth]{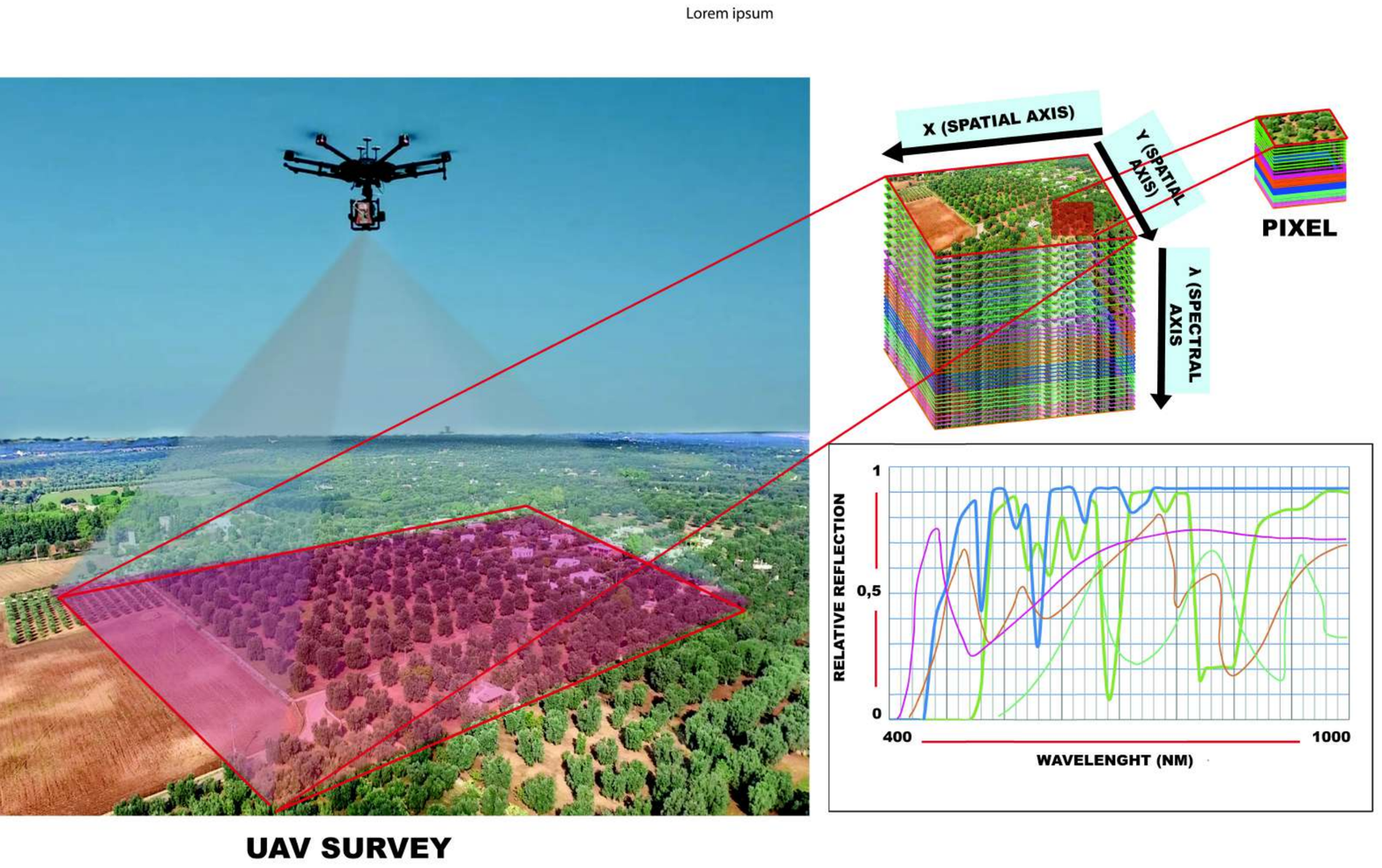}
        \caption{Hyperspectral Image acquisition, datacube and spectral content of several pixels (each color in the 2D plot represents a single pixel content).}
        \label{fig:hyperimage}
        
    \end{figure}
-\end{center}

HSI can be used in precision agriculture to assess the health of crops based on their distinctive signatures at different growth stages. The spectral behavior of a scene is recorded using HSI sensors, which are space-sensing devices that take many digital images of the same scene at once, each reflecting a contained or continuous spectrum. When a specific substance is subjected to a light source with a given spectral bandwidth, specific sections of the light are emitted, absorbed, and/or reflected depending on the substance's structure. The spectral signature of a material is the term used to describe this reaction \cite{manolakis2016hyperspectral}. Even though the hyperspectral image's data volume is always very high and has colinearity problems, chemometric methods are needed to extract the crucial intimate analysis. HSI can be used in precision agriculture to assess the health of crops based on their distinctive signatures at different growth stages. Space-sensing devices are called hyperspectral image sensors to capture the spectral behavior of a scene as a series of simultaneous digital images, each representing a contained or continuous spectral spectrum. When a specific substance is subjected to a light source with a given spectral bandwidth, specific sections of the light are emitted, absorbed, and/or reflected depending on the substance's structure. This reaction is referred to as the spectral signature of a material \cite{manolakis2016hyperspectral}. This information is stored in a cubic data structure, as seen in Fig \ref{fig:hyperimage}, where each spectral band is "stacked" by its wavelength. Therefore, the measurements of spectral responses enable the classification of distinct materials or the observation of specific compositional characteristics in biological subjects. Even though the hyperspectral image's data volume is always very large and has colinearity problems, chemometric methods are needed to extract the crucial intimate analysis.

\begin{figure}[H]
    \begin{center}
  \includegraphics[width=0.7\linewidth]{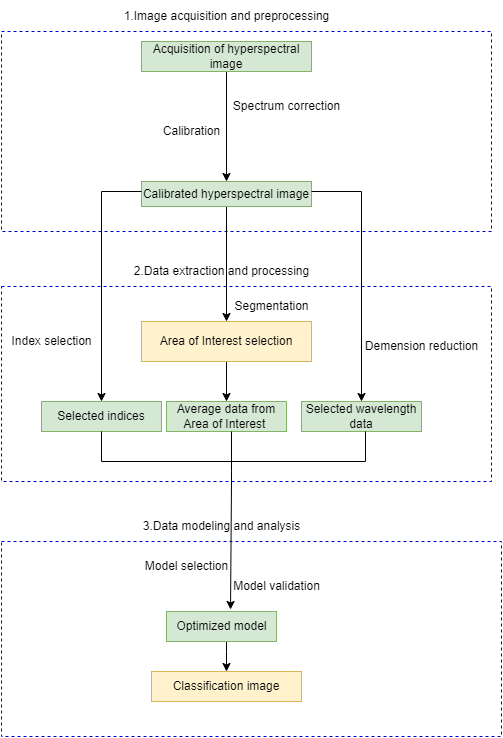}
  \caption{Flowchart of the typical steps for analyzing hyperspectral image data.}
  \label{fig:typicalsteps}
  \end{center}
  
\end{figure}

The acquisition of high-quality images that satisfy the study objectives is a crucial first step in the analysis of HSI. For accurate results, the proper selection of sensors and platforms is necessary, as well as the optimal spectral and spatial resolution settings, illumination design, scan rate, frame rate, and exposure time \cite{wu2013advanced}. The following stage is image pre-processing, which includes spectrum correction and calibration. The procedure consists of Standardizing the spectrum and spatial axes of the hyperspectral image, assessing the precision and reproducibility of the acquired data under various operating conditions, and removing instrumental errors and the curvature effect \cite{vidal2012pre}. Image segmentation is usually carried out as a preprocessing step before the formal spectral analysis to extract target objects from the background or create a mask that defines the area of interest for further extraction of information \cite{li2019detection}. The last step is selecting a model and applying it to the data, these can be regression models or classification models, depending on the goals of the research.

\section{Acquisition Modes}\label{acqmodes}

\begin{figure}[H]
    \begin{center}
  \includegraphics[width=0.9\linewidth]{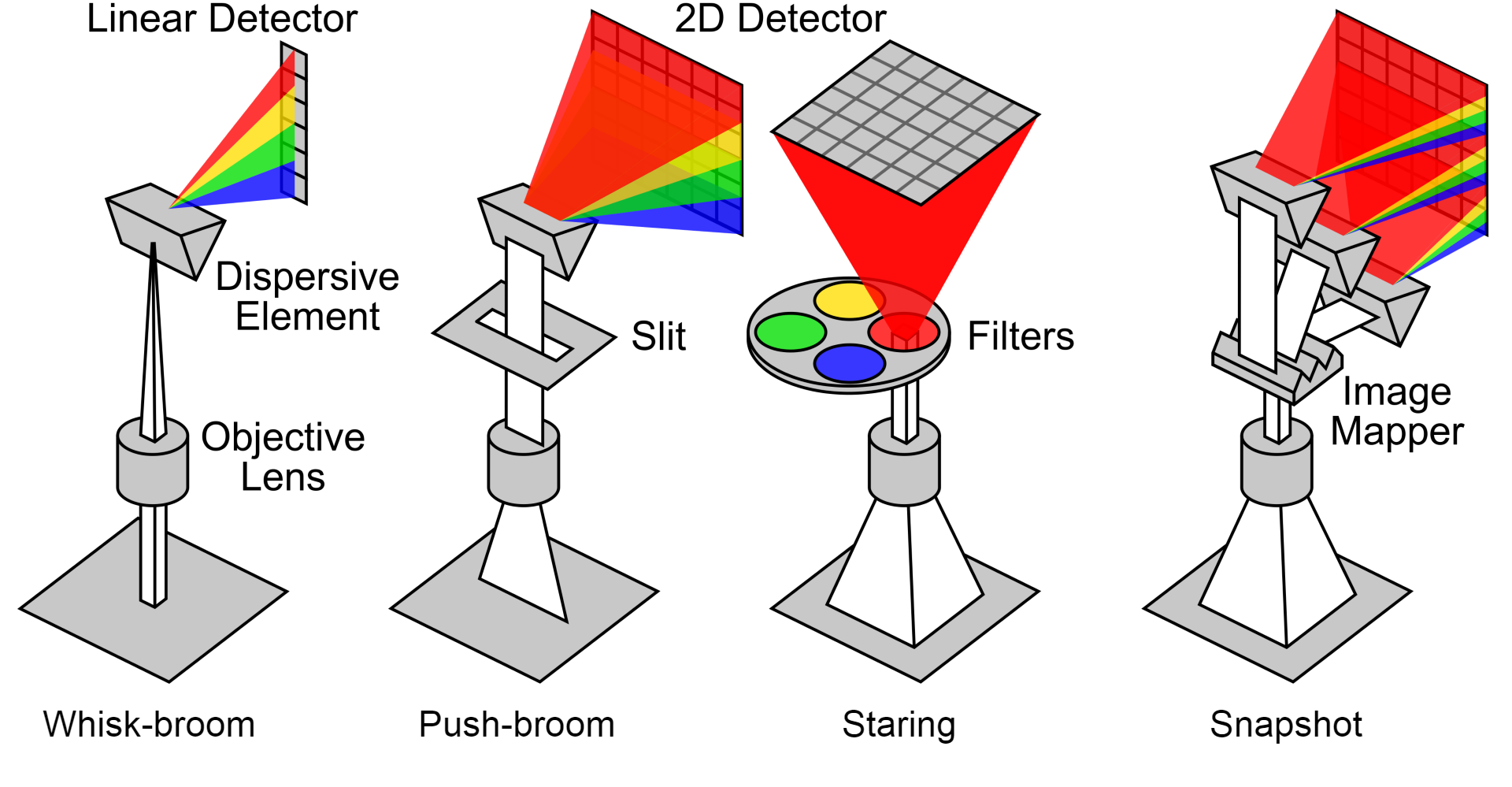}
  \caption{Imaging approaches}
  \label{fig:imgapproaches}
  \end{center}
  
\end{figure}

Based on the methods used to acquire both spectral and spatial information, hyperspectral systems are classified into 4 categories: whisk-broom, push broom, staring, and snapshot, in Fig \ref{fig:imgapproaches}.

\begin{enumerate}

    \item \textbf{whisk-broom}

    The whiskbroom imaging method is a technique used to acquire images in remote sensing applications. It involves scanning a target area or scene one line at a time, using a sensor that moves back and forth across the target area. In the whiskbroom imaging method, the sensor is typically mounted on a platform, such as an airplane or satellite, and moves across the target area in a series of parallel lines. As the sensor moves across the target area, it captures a series of narrow strips of the scene, one line at a time. These strips are then combined to create a complete image of the target area. In HSI, the sensor captures a series of images at different spectral wavelengths, which are combined to create a 3D hyperspectral image cube. The whiskbroom imaging method can be used to capture each of the individual images in the hyperspectral cube, providing a complete image of the scene at each spectral wavelength. One of the advantages of the whiskbroom imaging method is its ability to capture images with high spatial resolution. Because the sensor moves across the target area in a series of parallel lines, it can capture a large number of closely spaced image strips, which can be combined to create a high-resolution image of the target area. The whiskbroom imaging method is widely used in a variety of remote sensing applications, including environmental monitoring \cite{merlaud2013small, merlaud2018small}, mineral exploration \cite{beiranvand2014aster}, and defense and surveillance \cite{lee2003study}. It is particularly useful in applications where high-resolution images of the target area are required, or where rapid image capture is necessary. Overall, the whiskbroom imaging method is a powerful imaging technique that can be used in a variety of remote sensing applications. Its ability to capture high-resolution images of the target area makes it a useful tool for a wide range of applications.

    \item \textbf{Push broom} 

    The push broom method is a technique used to acquire hyperspectral images, particularly in remote sensing applications. It involves scanning a scene or target area one line at a time, using a hyperspectral camera that remains stationary \cite{mahlein2012hyperspectral}. The camera captures a series of images at different spectral wavelengths, which are combined to create a 3D hyperspectral image cube. The push broom method typically uses a line array sensor, which consists of a linear array of pixels that captures an image one line at a time. The sensor is mounted on a platform or satellite, which moves relative to the scene being imaged. As the platform moves, the sensor captures a series of images at different spectral wavelengths, one line at a time. The process is repeated until the entire target area has been scanned. Each line of the hyperspectral image cube represents the spectral information for each pixel in that line, across all the spectral wavelengths captured by the sensor. The resulting hyperspectral image cube contains information about the reflectance or absorption of each pixel at each wavelength, allowing for detailed analysis and interpretation of the data.
    The push broom method is preferred over other methods such as the whiskbroom method, which scans the scene using a scanning mirror or rotating prism to acquire an image line by line. The push broom method is generally considered to be more efficient and faster, as it requires only a linear array of pixels and does not require any mechanical components to move the sensor. The push broom method is widely used in a variety of remote sensing applications. It is particularly useful for monitoring large areas over time, as it allows for the detection of subtle changes in spectral signatures that can indicate changes in vegetation health \cite{hernandez2019early}, mineral content \cite{manley2014near}, or environmental conditions \cite{thomas2018quantitative}.

    \item \textbf{Staring}
    
    In the staring imaging method, the sensor captures a complete image of the target area or scene all at once. The sensor can be a camera or other imaging device that uses an array of pixels to capture the image. The sensor is typically mounted on a platform or satellite, which remains stationary during the image capture process \cite{guo2013novel}. Unlike the push broom method, which scans the scene one line at a time, the staring imaging method captures the entire scene at once. This can be useful in applications where a complete image of the scene is required, such as in surveillance or mapping applications \cite{zhang2017application}. The staring imaging method is typically used in conjunction with other techniques, such as HSI, to capture images with high spatial and spectral resolution. In hyperspectral imaging, the sensor captures a series of images at different spectral wavelengths, which are combined to create a 3D hyperspectral image cube. The staring imaging method can be used to capture each of the individual images in the hyperspectral cube, providing a complete image of the scene at each spectral wavelength. The staring imaging method is particularly useful in applications where a high-resolution image of the entire scene is required, or where rapid image capture is necessary. Overall, the staring imaging method is a powerful imaging technique that can be used in a variety of remote sensing applications. Its ability to capture a complete image of the scene in a single snapshot makes it a useful tool for a wide range of applications.

    \item \textbf{Snapshot}

    The snapshot imaging method is a technique used to acquire images in a single snapshot, particularly in optical imaging applications. It involves capturing an image of the entire field of view all at once, using a sensor that is designed to capture a large area in a single exposure \cite{johnson2007snapshot}. The snapshot imaging method typically uses a specialized sensor known as a focal plane array (FPA) \cite{arslan2015extended} or a detector array. The FPA is an array of photodetectors that captures the image of the entire field of view simultaneously. The FPA can be made up of different types of photodetectors, such as charge-coupled devices (CCDs) or complementary metal-oxide-semiconductor (CMOS) sensors, depending on the application. One of the main advantages of the snapshot imaging method is its ability to capture high-speed images of fast-moving objects or events. Because the entire field of view is captured in a single snapshot, there is no need for any mechanical scanning or motion of the sensor, which allows for rapid image capture. The snapshot imaging method is widely used in a variety of applications, including high-speed imaging, microscopy \cite{gao2010snapshot}, astronomy \cite{offringa2014wsclean}, and biomedical imaging \cite{yu2021microlens}. In high-speed imaging applications, the method can be used to capture images of fast-moving objects or events, such as explosions or high-speed collisions. In microscopy, the method can be used to capture images of small, fast-moving particles or cells. In astronomy, the method can be used to capture images of distant stars and galaxies. Overall, the snapshot imaging method is a powerful imaging technique that can be used in a variety of optical imaging applications. Its ability to capture high-speed images of fast-moving objects and events makes it a useful tool in many scientific and industrial applications.
\end{enumerate}

\section{Datasets for HSI in Agriculture}\label{datasets}

With the development of AI, data and performance measures have assumed an ever-larger significance. Recently, a few new HSI datasets have been available for the training and validation of deep learning networks. We initially explain various performance criteria for the evaluation and comparison of various algorithms before introducing accessible open-source datasets that support the research.

\subsection{Open-source Datasets}

Table \ref{Tab:datasets} displays a few of the suggested HSI datasets for agriculture. More HSI data can be gathered as this technique develops, allowing for the availability of larger datasets. The quantity of data, spatial resolution, spectral channels, and variety of scenarios are the most critical characteristics of the available datasets.

\begin{table*}
\centering
\begin{minipage}{\textwidth}
\caption{Hyperspectral agricultural datasets}\label{Tab:datasets}
\begin{adjustbox}{width=\textwidth}

\begin{tabular}{ p{0.2\textwidth}  p{0.082\textwidth}  p{0.2\textwidth}  p{0.15\textwidth}  p{0.082\textwidth}  p{0.12\textwidth}  p{0.09\textwidth}  p{0.082\textwidth}  p{0.082\textwidth} }
 \hline

Dataset & Year & Source & SD(pixels) & SB & WL(nm) & 
Samples & Classes & SR(m) \\ [0.5cm] 
 \hline
 Indian Pines \cite{scenes2020available} & 1992 & NASA AVIRIS & 145 x 145 & 220 & 400 - 2500 & 10249 & 16  & 20  \\ [0.5cm]
 \hline
 
  Salinas \cite{scenes2020available} & 1998 & NASA AVIRIS & 512 x 217 & 224 & 360 - 2500 & 54129 & 16  & 3.7  \\ [0.5cm]
 \hline
 
 Pavia University \cite{scenes2020available} & 2001 & ROSIS-03 sensor & 610 x 610 & 115 & 430 - 860 & 42776 & 9 & 1.3 \\ [0.5cm]
 \hline

 Botswana \cite{scenes2020available} & 2004 & NASA EO-1 & 1496 × 256 & 242 & 400-2500 & 3248 & 14  & 30 \\ [0.5cm]
 \hline

  Chikusei \cite{yokoya2016airborne} & 2014 & Headwall Hyperspec- \newline VNIR-C imaging sensor & 2517x2335 & 128 & 363-1018 & 77592 & 19  & 2.5 \\ [0.5cm]
 \hline
 
 WHU-Hi-HanChuan \cite{zhong2018mini} & 2016 & Headwall Nano-Hyperspec \newline imaging sensor & 1217 × 303  & 274 & 400-1000 & 257530 & 16  & 0.109 \\ [0.5cm]
 \hline
 
 WHU-Hi-HongHu \cite{zhong2018mini} & 2017 & Headwall Nano-Hyperspec \newline imaging sensor & 940 × 475 & 270 & 400-1000 & 386693 & 22  & 0.043 \\ [0.5cm]
 \hline

  WHU-Hi-LongKou \cite{zhong2018mini} & 2018 & Headwall Nano-Hyperspec \newline imaging sensor & 550 × 400 & 270 & 400-1000 & 204542 & 9 & 0.463 \\ [0.5cm]
 \hline

\end{tabular}
\end{adjustbox}
\end{minipage}
\end{table*}

Numerous papers used Indian pines \cite{scenes2020available} and University of Pavia datasets \cite{scenes2020available}. Both datasets are captured by airborne hyperspectral-imaging sensors and contain pixel-level ground truth. The Indian Pines dataset, which comprises 224 band hyperspectral images, was captured by the AVIRIS sensor to target LULC in the agricultural domain. Researchers frequently utilize this dataset to analyze LULC patterns, and typically focus on the 200 spectral bands while excluding the water absorption bands. The Salinas dataset, which focuses on different agricultural classes, is captured by the same sensor used to capture the Indian Pines dataset. These two datasets are quite similar in terms of their data types. The ROSIS airborne sensor captured the University of Pavia dataset, producing pictures with 103 spectral bands. Botswana \cite{scenes2020available} is another airborne pixel-level labeled imagery dataset used for land cover classification. The Wuhan University RSIDEA research group has gathered and made available the WHU-Hi dataset (Wuhan UAV-borne hyperspectral image) \cite{zhong2018mini}, which is used as a benchmark dataset for studies on accurate crop classification and hyperspectral image classification. Three distinct UAV-borne hyperspectral datasets are included in the WHU-Hi dataset: WHU-Hi-LongKou, WHU-Hi-HanChuan, and WHU-Hi-HongHu. All of the data were collected in Hubei province, China, in agricultural areas that grew a variety of crops. The spectral classes from the Pavia University and Salinas datasets are homogeneously distributed throughout the hyperspectral image \cite{yadav2019study}. WHU-Hi-LongKou and Pavia University datasets have a lesser number of classes compared with other datasets. Unmanned aerial vehicle (UAV)-borne hyperspectral systems using Headwall Nano-Hyperspec sensor can acquire hyperspectral imagery with a high spatial resolution \cite{zhong2020whu}. In the Chikusei dataset \cite{yokoya2016airborne}, the Hyperspec-VNIR-C imaging sensor was used to capture hyperspectral data over urban and rural regions in Chikusei, Ibaraki, Japan. The dataset contains ground truth high-resolution color images captured by EOS 5D Mark II for 19 classes.

\section{Deep Learning-Based Approaches for HSI Classification}\label{techniques}

\begin{center}
    \begin{figure}[H]
  \includegraphics[width=\linewidth]{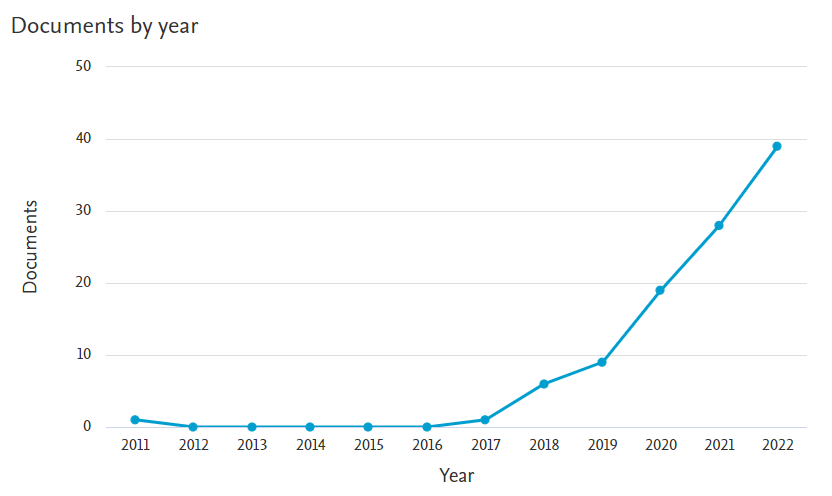}
  \caption{Number of published articles by year on deep learning with hyperspectral data applied in agriculture sector, (source: Scopus).}
  \label{fig:DLarticlesGraph}
  
\end{figure}
\end{center}

\begin{center}
    \begin{figure}[H]
  \includegraphics[width=\linewidth]{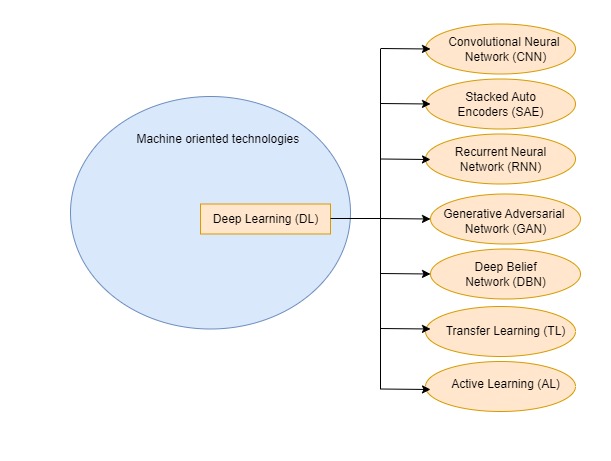}
  \caption{The various categories of prominent deep learning techniques utilized for HSI classification}
  \label{fig:DLtechniques}
  
\end{figure}
\end{center}

The accuracy of ML technologies is rising steeply because of their built-in mechanical capabilities such as feature extraction, selection, and reduction of spatial-spectral and contextual features. Not only are these technologies intelligent and cognitive, but they also possess a high degree of precision \cite{hassanzadeh2020yield}. The most recent Deep Learning (DL) techniques for classifying hyperspectral data, including CNN, SAE, RNN, GAN, DBN, TL, and AL, are presented in Figure \ref{fig:DLtechniques} and elaborated upon in detail below. Table \ref{DLComp} illustrates some of the DL classification methods. According to the Scopus statistics, there are 109 relevant papers from 2011 to 2023 where "hyperspectral images" and "agriculture" and "deep learning" are used as keywords Fig. \ref{fig:DLarticlesGraph}. It is interesting to see how in 2018 there has been a strong increase of the published papers in the agriculture sector, thanks to a better availability of the deep learning frameworks.

\begin{center}
    \begin{figure}[H]
  \includegraphics[width=\linewidth]{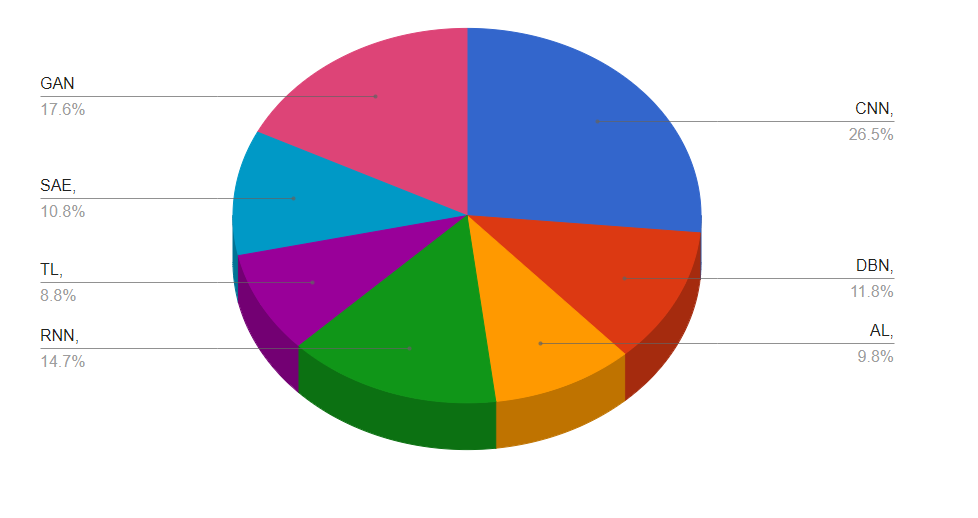}
  \caption{Pie-chart of related articles on DL approaches used for HSI classification, (source: Scopus).}
  \label{fig:DLarticlespie}
  
\end{figure}
\end{center}

The distribution of literary studies analyzed for each of the selected DL techniques is shown in Figure \ref{fig:DLarticlespie} as a pie-chart with percentage values for each category. 

\subsection{Convolutional neural network (CNN)}

\begin{center}
    \begin{figure}[H]
  \includegraphics[width=\linewidth]{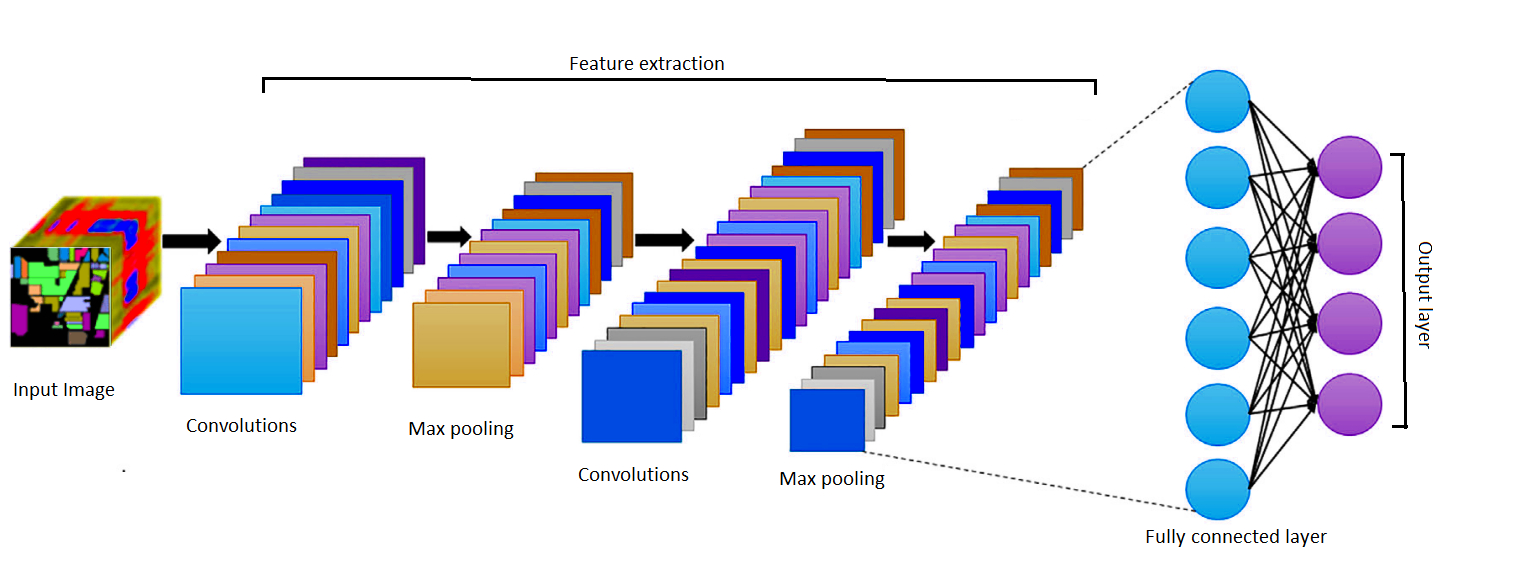}
  \caption{Graphical representation of CNN}
  \label{fig:CNNarch}
  
\end{figure}
\end{center}

The most widely used neural network for classifying images is the convolutional neural network (CNN), whose primary structural unit is the convolutional (CONV) layer. In comparison to other methods, CNN has been widely employed for image classification \cite{blaschke2014geographic}, detection \cite{tao2016robust} and segmentation \cite{bougourzi2023pdatt, bougourzi2023d}. Deep neural networks are capable of learning deep feature representation for analyzing hyperspectral images and can achieve excellent classification accuracy in various datasets. CNN has become a popular technique for LULC classification due to its exceptional ability to process hyperspectral images effectively by extracting spectral-spatial discriminative features, which is evident in numerous studies \cite{lin2016land, wambugu2021hybrid}. In fact, several studies have shown that CNN outperforms traditional machine learning algorithms such as Random Forest, Support Vector Machine, and k-Nearest Neighbors on multiple datasets \cite{carranza2019framework}, authors have proposed a per-superpixel model that combines multi-scale CNN for LULC classification and leverages high-level feature extraction. This per-superpixel multi-scale CNN approach has effectively addressed misclassification caused by the scale effect in complex LULC classes, surpassing the per-superpixel single-scale CNN method, as evidenced by the results of \cite{chen2019superpixel}. Another notable advancement in this field is the introduction of a deep hybrid dilated residual network (DHDRN) in \cite{cao2020deep}, which has been evaluated on three publicly available hyperspectral datasets and compared to state-of-the-art methods. The experimental findings have highlighted that the proposed DHDRN method achieves superior classification accuracy and efficiency, surpassing previous methods. Specifically, the DHDRN method has achieved impressive overall classification accuracy rates of 99.26\%, 99.44\%, and 97.96\% on the Pavia University, Indian Pines, and Salinas datasets, respectively. The authors have conducted ablation studies to understand the contribution of each component of the proposed method, which has revealed that the hybrid architecture and dilated convolutional layers play a crucial role in achieving the best classification performance. The Deep Residual Network (ResNet) \cite{he2016deep} is considered a significant milestone in the history of CNN. ResNet has addressed the problem of training deep CNN models \cite{7780459}. Recently, ResNet has been successfully used in hyperspectral image analysis, such as hyperspectral image classification \cite{8127330}, hyperspectral image denoising \cite{yuan2018hyperspectral}, increasing the spatial resolution of hyperspectral images \cite{wang2017deep}, and unsupervised spectral-spatial feature learning of hyperspectral images \cite{mou2017unsupervised}. Fig \ref{fig:CNNarch} shows a graphical representation of CNN architecture.

\subsection{Stacked auto‑encoder (SAE)}

\begin{center}
    \begin{figure}[H]
  \includegraphics[width=\linewidth]{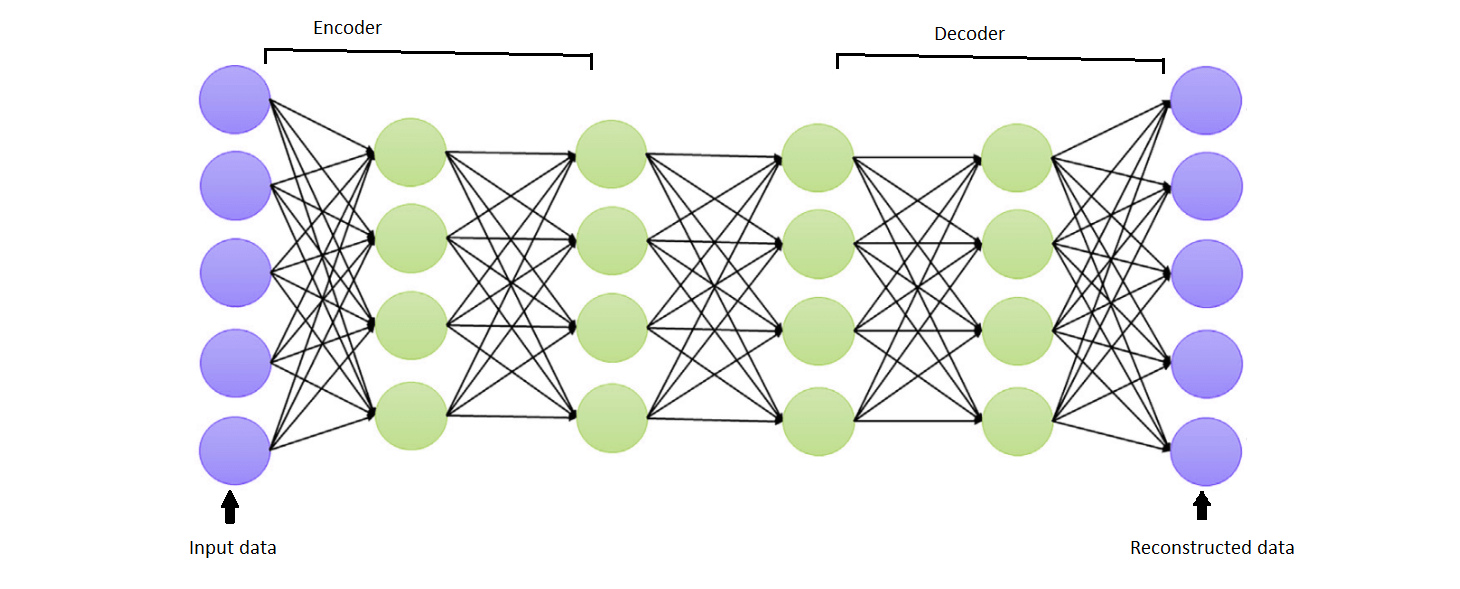}
  \caption{Diagrammatic portrayal of SAE}
  \label{fig:SAEarch}
  
\end{figure}
\end{center}

In 2016, \cite{xing2016stacked} utilized the SAE model to obtain practical high-level features for remote sensing image classification. This was the first application of deep learning to HSI analysis. The input data was reconstructed using auto-encoders (AE) which are composed of the encoder and decoder. The AE was trained separately and connected to each layer of the SAE \cite{bengio2006greedy}. The abstract features were extracted by rebuilding the input data layer by layer. During the unsupervised pretraining stage, the features learned from one AE were used as input data for training the next AE in a greedy way, thus reducing each AE's reconstruction error. After pretraining, the parameters, such as weights and biases, of all AEs were used as initial values for SAE. The parameters of each layer were adjusted using backpropagation of error when the labeled data was used as the supervised signal, and the parameters of the structure were updated using the stochastic gradient descent algorithm. \cite{vincent2008extracting, vincent2010stacked} proposed denoising auto-encoders (DAE) and stacked denoising auto-encoders (SDAE) as other enhancement strategies. Authors in \cite{zhou2019learning} presented a novel and robust approach for hyperspectral image classification using a compact and discriminative stacked autoencoder (CDSAE). The approach consisted of two steps, in the initial step, low-dimensional discriminative features were extracted by applying a local Fisher discriminant regularization to each hidden layer of the SAE. In the second step, an effective classifier was integrated into the training process. The proposed method was evaluated on three different HSI datasets, and the results demonstrated its remarkable superiority in accurate and reliable hyperspectral image classification. Fig \ref{fig:SAEarch} depicted a graphical illustration of SAE.

\subsection{Deep belief network (DBN)}

\begin{center}
    \begin{figure}[H]
  \includegraphics[width=1.2\linewidth]{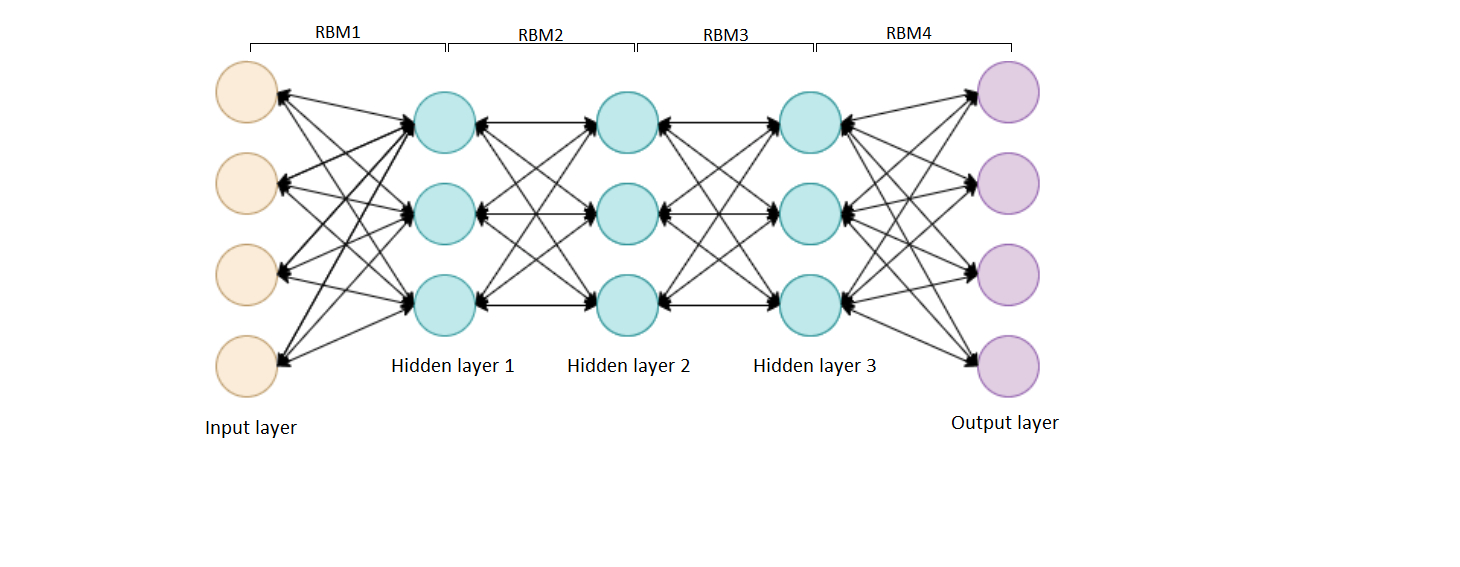}
  \caption{Diagrammatic illustration of DBN}
  \label{fig:DBNarch}
  
\end{figure}
\end{center}

In 2006, Hinton proposed the deep belief network (DBN) \cite{hinton2006fast}, which uses the Restricted Boltzmann Machine (RBM) as a learning module similar to SAE's use of auto-encoders. However, DBN employs a symmetrical connection structure, consisting of several RBMs, with connections between the layers rather than within the units of each layer. The output of one layer serves as the input for the next layer. The RBM layers are initially pre-trained in an unsupervised manner using unlabeled samples to preserve the characteristics as much as possible. The entire DBN network is fine-tuned using a small number of labeled samples and the backpropagation algorithm \cite{chen2015spectral}. The deep features extracted are used for detection and classification tasks. Random initialization of weight parameters is critical because it enables DBN networks to overcome the primary limitations of the backpropagation technique, such as local optimization and long training times. In \cite{gavade2020hybrid}, the authors utilized the Firefly Harmony Search Deep Belief Network (FHS-DBN) model for LULC classification on four benchmark datasets. The DBN was trained using a hybrid approach combining the Firefly Algorithm (FA) and the Harmony Search (HS) algorithm to obtain the FHS algorithm. This approach showed promising results for LULC classification, demonstrating the potential of integrating multiple optimization algorithms in deep learning models for improved performance. As DBNs have been used in hyperspectral image classification for some time, their efficacy in the field is well-established. While DBNs have been shown to be effective for a wide range of tasks, they also have some challenges, including:

\begin{itemize}
    \item Lack of Transferability: DBNs are often trained on one specific dataset and are not easily transferable to other HSI datasets, which can limit their ability to generalize to new hyperspectral images
    \item Over-complexity: DBNs can have a large number of parameters, which can lead to over-complex models that are difficult to train and may not generalize well to new hyperspectral images
    \item Limited Ability to Handle Noisy Data: DBNs can be sensitive to noisy data, which is a common problem in HSI due to various environmental factors such as atmospheric turbulence, clouds, and shadows
    \item High Computational Cost: DBNs can be computationally expensive to train, especially for large hyperspectral images, which can limit their applicability in real-world scenarios
\end{itemize}
Fig \ref{fig:DBNarch} represents schematic illustration of the DBN.

\subsection{Recurrent neural networks (RNN)}
The concept of a recurrent neural network (RNN) was first introduced by Williams in 1989 \cite{williams1989learning}. RNNs differ from feed-forward neural networks by incorporating a recurrent hidden state that depends on previous steps. This enables RNNs to recognize patterns in data sequences and temporal properties. Recently, RNNs have been used to classify hyperspectral images, as they can efficiently analyze hyperspectral pixels as sequential data \cite{mou2017deep}. However, standard RNN models suffer from issues such as gradient explosion or disappearance, which have been partially resolved with the introduction of Long Short-Term Memory Networks (LSTMs) \cite{jiang2018lstm} and Gated Recurrent Units (GRUs) \cite{chung2014empirical}. Different LSTM models, such as LSTM-F, LSTM-S (unidirectional), and base LSTM (b-LSTM)-S (bidirectional), have been introduced to address sliding-window segmentation and operate in various modes. The bidirectional-convolutional long and short-term memory network (Bi-CLSTM) has been used to learn spectral-spatial characteristics automatically from hyperspectral data, resulting in improved classification performance of about 1.5\% compared to a 3D-CNN \cite{liu2017bidirectional}. A  method for HSI classification, called Spectral-Spatial LSTM (SS-LSTM), has been proposed in \cite{zhou2019hyperspectral}. This method utilizes Convolutional Neural Networks (CNNs) and Principal Component Analysis (PCA) to extract spectral and spatial features from the hyperspectral image, respectively. These features are then fed into separate LSTM layers to capture the temporal dependencies and interdependencies between them. The SS-LSTM is trained on a large hyperspectral dataset and evaluated on three benchmark datasets.

\begin{center}
    \begin{figure}[H]
  \includegraphics[width=\linewidth]{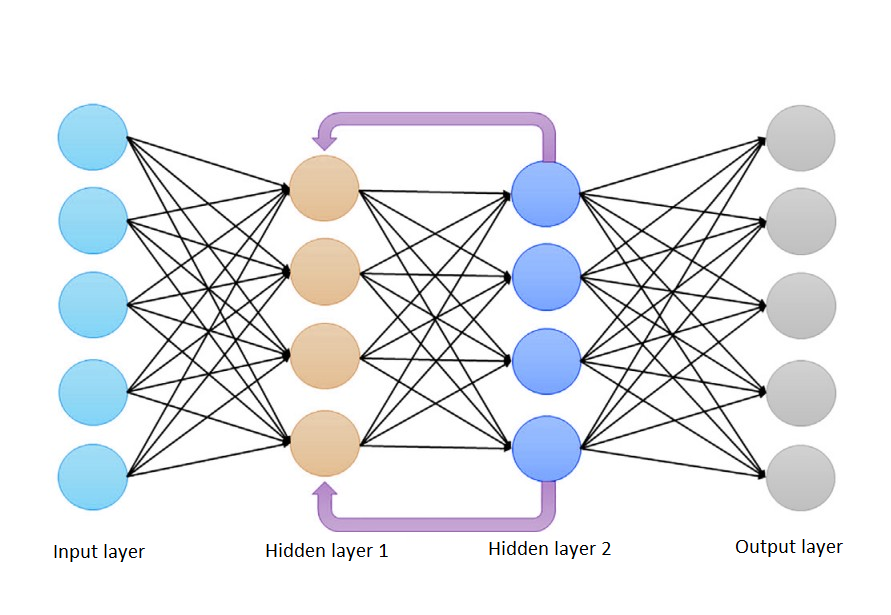}
  \caption{Graphical representation of RNN}
  \label{fig:RNNarch}
  
\end{figure}
\end{center}

There are several problems associated with applying Recurrent Neural Networks (RNNs) to HSI data : 
\begin{itemize}
    \item Handling Structural Changes: RNNs can have difficulty in handling structural changes in hyperspectral images, such as changes in atmospheric conditions, illumination, and viewpoint, which can impact their performance on hyperspectral image classification tasks
    \item Data Variability: Agricultural hyperspectral images can vary significantly due to changes in weather conditions, soil conditions, and plant growth stages, making it challenging to train RNNs effectively
    \item Training Difficulties: Training RNNs on HSI data can be challenging due to the sequential nature of the data, and the need for appropriate training algorithms and techniques to handle the data effectively
\end{itemize}
Fig \ref{fig:RNNarch} represents a graphical representation of RNN.

\subsection{Generative adversarial networks (GAN)}
\begin{center}
    \begin{figure}[H]
    \centering
  \includegraphics[width=0.4\linewidth]{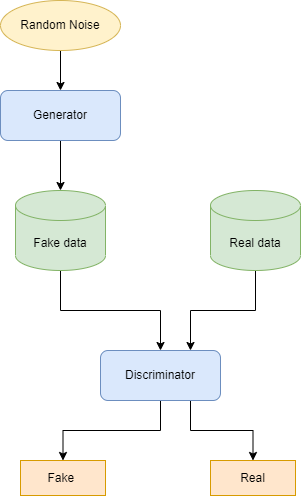}
  \caption{Schematic representation of GAN}
  \label{fig:GANarch}
  
\end{figure}
\end{center}

Generative Adversarial Network (GAN) was first proposed in 2014, which generates samples based on required class label through adversarial training \cite{goodfellow2014generative}. Generative techniques aim to identify the distribution parameters from the data and generate new samples according to the identified models. Several improved GANs, including Deep Convolutional GAN \cite{chen2019hyperspectral}, 1-D and 3-D GAN \cite{zhu2018generative}, Capsule GAN \cite{xue2020general}, Cascade Conditional GAN \cite{liu2020cascade}, MDGAN \cite{gao2019hyperspectral}, and 3DBF-GAN \cite{he2017generative}, have been utilized for hyperspectral imaging. GANs have shown very promising results with a small number of labeled samples by fully exploiting sufficient unlabeled samples \cite{zhan2017semisupervised}. There are several challenges in applying Generative Adversarial Networks (GANs) to HSI data, including:
\begin{itemize}
    \item Data heterogeneity: hyperspectral data can have heterogeneous features, making it difficult for GANs to capture the full range of information present in the data
    \item Limited labeled data: In many applications of HSI, labeled data is limited, which can impact the ability of GANs to learn effectively from the data
    \item Data variability: hyperspectral data can be affected by various environmental factors, such as atmospheric conditions, which can lead to significant variability in the data
    \item Data imbalance: hyperspectral data often has imbalanced class distributions, which can affect the performance of GANs
\end{itemize}
Fig \ref{fig:GANarch} represents schematic illustration of the GAN.

\subsection{Active learning (AL)}*

Active learning (AL) is a technique that shows promise in addressing the challenge of limited labeled samples in HSI. It involves an iterative process of selecting the most informative examples from a subset of unlabeled samples based on their uncertainty and intrinsic distribution and structure \cite{cui2019double,lei2021active}. AL is more efficient than traditional semi-supervised learning methods and can train deep networks with fewer training samples \cite{bhardwaj2020spectral}. Various AL approaches have been proposed for HSI classification, including random sampling (RS) \cite{smith2018less}, maximum uncertainty sampling (MUS) \cite{nguyen2022measure}, multiview (MV) \cite{zhang2019adaptive}, and mutual information (MI)-based sampling \cite{liu2017feature}. Applying AL approaches to HSI data presents several challenges, including:


\begin{itemize}
    \item Query strategy selection: Selecting the appropriate query strategy is crucial for effective active learning in HSI. There is a trade-off between the cost of acquiring labels and the quality of the model being learned, and different query strategies may perform better or worse depending on the specifics of the data and task at hand
    \item Data variability: Agricultural data can be affected by various environmental factors, such as weather conditions and soil variability, which can lead to significant variability in the data. This variability can make it difficult for active learning algorithms to accurately model the underlying data distribution
    \item Label noise: Labeling data in agriculture can be subjective and prone to human error, leading to label noise in the data. This can impact the performance of active learning algorithms that rely on accurate labels
\end{itemize}

\subsection{Transfer learning (TL)}

The application of transfer learning models to HSI analysis has proved to be successful and reliable. A few top layers of CNN are developed using a small number of training samples, while the bottom and middle layers can be transferred from models of other scenarios \cite{yang2016hyperspectral}. The overall classification accuracy of CNN-transfer is higher than CNN when training samples are low. Hyperspectral image super-resolution is also a challenge.In order to improve the resolution of hyperspectral images, a novel framework is developed that makes use of information from natural images \cite{yuan2017hyperspectral}. The proposed approach utilizes transfer learning to extend the mapping between low and high-resolution images, which is learned by a deep convolutional neural network, to the hyperspectral domain. There are several challenges in applying transfer learning to HSI data, including:

\begin{itemize}
    \item Domain shift: The distribution of the source and target data may differ significantly, causing a domain shift. This can impact the effectiveness of transfer learning algorithms
    \item Task specificity: The target task in agriculture may be different from the source task, which can impact the ability of transfer learning algorithms to effectively transfer knowledge
    \item Model compatibility: The source and target data may have different dimensionalities or structures, which can impact the ability of transfer learning algorithms to effectively transfer knowledge
\end{itemize}

\begin{table}[h!]
\begin{center}
\caption{Comparison analysis of DL Classification Techniques}\label{DLComp}
\begin{tabular}{M{0.5in}M{0.5in}M{2in}M{2in}}
\toprule%
Year & Article & Methodology & Results \\ \midrule

2016 & \cite{zabalza2016novel} & Bands segmentation using SSAE, local SAEs to process original features & 
\makecell{IP: OA-80.66\% \\ PU: OA-97.5\%} \\

2018 & \cite{singh2018efficient} & LPP, Deep features using SAE & \makecell{IP: OA-84.4\%. \\ SA: OA-87.2\%.} \\

2018 & \cite{paoletti2018new} & 3DCNN  & \makecell{IP: OA 98.37\%, AA 99.27\% \\PU: OA 98.06\%, AA 98.61\%.} \\

2019 & \cite{zhou2019hyperspectral} & Spectral-spatial LSTM, PCA & \makecell{IP: OA 95.00\%, k 94.29\%\\ PU: OA 98.48\%, 97.56\%\\ SA: OA 97.89\%, k 97.65\%} \\

2020 & \cite{vaddi2020hyperspectral} & Data normalization and CNN-based classification & \makecell{IP: OA 99.02\%, AA 99.17\% \\ PU: OA 99.94\%, AA 99.92\%}\\

2020 & \cite{cao2020deep} & 3D-2D spectral spatial hybrid dilated residual networks & \makecell{IP: OA 99.46\%, k 99.38\%\\ PU: OA 99.81\%, k 99.74\%}\\

2021 & \cite{singh2021pre} & SAE, PCA and LPP-based spectral-spatial features & \makecell{IP: OA 99.21\%, AA 99.19\% \\ PU: OA 99.89\%, AA 99.90\%}\\

2022 & \cite{zhan2022enhanced} & Combination of LSTM, residual network and spectral-spatial attention network &  \makecell{IP: OA 97.69\%, AA 97.19\% \\ PU: OA 95.87\%, AA 95.37\% \\ SA: OA 98.34\%, AA 98.84\%}\\

2022 & \cite{pande2022hyperloopnet} &  Self-looping CNN & \makecell{PU: OA 77.76\%, AA  71.36\%\\ SA: OA 99.32\%, AA 99.68\%} \\

2022 & \cite{yadav2022multi} &  Combination of differential texture pattern extraction and PCAL & \makecell{IP: OA 97.6\%\\ PU: OA 98.48\%} \\

2022 & \cite{li2022manifold} & Termed manifold-based multi-DBN (MMDBN) & \makecell{IP: OA 81.50\%\\ SA: OA 91.79\% \\ BO: OA 94.06\%} \\

2023 & \cite{alkhatib2023tri} & Multi-scale 3D-CNN and three-branch feature fusion & \makecell{PU: OA 92.66\%, AA  90.65\%\\ SA: OA 96.68\%, AA 97.69\%} \\

2023 & \cite{zhou2023shallow} & Shallow-to-Deep Feature Enhancement SDFE with CNN & \makecell{IP: OA 99.16\%, AA  99.07\%\\ PU: OA 99.80\%, AA 99.70\%, \\ SA: OA 99.97\%, AA 99.80\%} \\

\bottomrule
\end{tabular}
\end{center}
\end{table}

\section{Applications of HSI technology in agriculture}\label{apps}
This section illustrates and outlines the most significant contributions of different agricultural sectors in HSI.

\subsection{Soil analysis}
Crop growth requires a healthy soil environment. Quick and accurate access to information on soil nutrient content is a requirement for scientific manuring. Poor soil management threatens the quality and effectiveness of the soils, which are a major factor in the rural generation \cite{song2018predicting}. The various factors impacting soil and soil erosion, such as bright sun and heavy rain, are greatly influenced by the regional climate. In order to address agricultural issues, such as crop quality and yield, soil erosion must be identified. HSI technology can help in soil analysis by capturing a large amount of data across a wide range of the electromagnetic spectrum. The resulting images can then be analyzed to identify and quantify various soil properties, such as soil moisture content \cite{wu2012rapid}, nutrient levels \cite{jia2017hyperspectral}, and mineral composition \cite{manley2014near}. This information can be used to map soil characteristics and support more informed decision-making for agriculture, environmental management, and other soil-related applications. Researchers look for simpler, non-destructive methods to identify soil organic matter (SOM), as it is essential in the soil-plant ecosystem. Table \ref{soilan} summarizes some of the most important research on HSI for soil analysis applications.

\begin{table}
\begin{center}

\caption{An overview of the key research conducted on soil analysis using HSI}\label{soilan}

 \begin{tabular}{p{1.5in}p{0.5in}p{1.4in}p{2.2in}}
\toprule%
Objective & Article & Method & Results \\ \midrule

Measuring soil TAs content & \cite{wei2021estimating} & DNN-CARS & R2CV = 0.69, RMSECV = 0.61, RECV = 6.56 \\
  
The concentration of nitrogen in soil & \cite{patel2020deep} &  DASU DASU-based DL network & 46.6\% N for g/100 g sample\\ 
  
Prediction of soil organic carbon (SOC) content & \cite{hong2020exploring} & fractional order derivative (FOD), Random forest (RF) & highest R2CV = 0.66 \\ 
  
Prediction of heavy metal concentrations in agricultural soils & \cite{tan2020estimation} & RF, SRF, RRF, GRF, HySpex VNIR-1600 & Rp2 =0.75, RMSEp = 8.24\\ 
  
Detection of soil organic matter & \cite{reis2021detection} & Partial Least Squares Regression (PLSR) & R2 = 0.75, r = 0.87  , RPD = 2.1 \\ 
  
A model for estimating the content of Pb in soil & \cite{tian2020hyperspectral} & MLR, PCR & R2 =0.724\%, RMSE= 24.92\% MRE= 28.22\%\\

estimation of soil properties & \cite{singh2022quantitative} &  Hybrid features, LSTM & R2= 0.85, RMSE= 10.56 \\

Estimating the concentration of soil heavy metals & \cite{annam2023estimating} & VGG19 transfer learning & Model accuracy = 81.25\%, RMSEas=2.89, RMSEcd=0.12, RMSEpb=0.22\\
\bottomrule
\end{tabular}

\end{center}
\end{table}

\subsection{Crop yield estimation}
Yield prediction is one of the most significant areas of precision agriculture research. Crop management, crop supply matching with demand, yield prediction, yield mapping, and crop supply mapping are essential for maximizing production \cite{al2016prediction}. One of the major issues in agricultural management that can be solved most effectively by precision farming methods is crop production estimation. HSI technology can help with crop yield estimation by providing information about the health and vigor of crops, including data on plant chlorophyll content, water content, and nutrient levels. This information can be used to estimate crop yield, predict crop stress and potential yield losses, and optimize crop management practices, leading to improved yields. By capturing detailed spectral information from the visible and near-infrared regions of the electromagnetic spectrum, HSI can also identify and map specific plant species and vegetation types, and detect changes in the landscape over time, which are all valuable for crop yield estimation. Table \ref{cropyield} summarizes the most important research on HSI for crop yield estimation applications, 

\begin{table}
\begin{center}
\caption{An overview of the key research on the application of HSI for crop yield prediction.}\label{cropyield}

 \begin{tabular}{p{1.5in}p{0.5in}p{1.4in}p{2.2in}}
\toprule%
Objective & Article & Method & Results \\ \midrule

Predection of corn yield & \cite{yang2021estimation} &  2DCNN & Kappa coefficient K=0.69, classification accuracy=75.50\% \\
 
Forecasting carrot crop yield & \cite{suarez2020accuracy} &  WV2 & R2<0.57, p<0.05\\
 
Prediction of Corn Seeds & \cite{pang2020rapid} & CNN(1D and 2D)+ HSI & Recognition accuracy =90.11\% (1DCNN) Recognition accuracy =99.96\% (2DCNN)\\
  
Forage yield and quality estimation & \cite{geipel2021forage} & PPLSR, neutral detergent fiber (NDF), SLR & RMSE Fresh matter = 14.2\%. RMSE Dry matter = 15.2\% RMSE Protein Content = 11.7\%\\

Classify of different crop types & \cite{farmonov2023crop} & WA-CNN & OA = 97.89\% and user accuracy from 97\% to 99\% \\

\bottomrule
\end{tabular}
\end{center}
\end{table}

\subsection{Agricultural crop classification }

One significant area of study for several agricultural applications is the identification and classification of the crop using hyperspectral images. HSI technology can assist in agricultural crop classification by capturing detailed spectral information of crops in the visible and near-infrared regions of the electromagnetic spectrum. This information can be used to identify unique spectral signatures for different crop types and growth stages, which can then be used for classification purposes. Machine learning algorithms are typically employed to analyze the hyperspectral data and classify crops based on their spectral reflectance characteristics. This information can be used to create maps of crop types and growth stages, which can provide valuable insights for agricultural management and decision-making. Additionally, HSI can be used to detect crop stress \cite{romer2012early}, diagnose potential yield-limiting factors \cite{vigneau2011potential}, and monitor crop health, which are all important for crop management and optimization. HSI can also be used to monitor changes in the landscape over time, allowing for the detection of crop growth and yield changes and providing a useful tool for precision agriculture. Table \ref{agricanalysis} summarizes some of the most important research on HSI for agricultural crop classification applications.
\begin{table}
\begin{center}

\caption{An overview of the most significant studies conducted on agricultural crop classification using HSI}\label{agricanalysis}
 \begin{tabular}{p{1.5in}p{0.5in}p{1.4in}p{2.2in}}
\toprule%
Objective & Article & Method & Results \\ \midrule

Corn seed variety classification & \cite{zhang2021corn} &  DCNN, KNN & DCNN training accuracy=100\% Testing accuracy rate=94.4\%, 57 and validation accuracy rate= 93.3\%\\ 
  
Variety identification of coated maize kernels & \cite{zhang2020application} & LR, SVM, CNN,RNN and LSTM & Classification accuracy over 90\%\\  
  
Prediction of intact oranges consistency parameters & \cite{riccioli2021optimizing} & ANN & (SSC)RMSECV =0.87\% (TA) RMSECV =0.23  RMSECV =2.78 for MI  RMSECV =1.11 for brima\\ 
  
The inherent uniformity of apple fruit slices is accentuated & \cite{lan2021method} & Near-infrared HSI & Rcv2=0.83, TSC Rcv2=0.81, RPD=2.20 and RPD=2.39 \\ 

Predict the viability of pepper seeds & \cite{hong2023nondestructive} & HSI PLS-SVM, Xray Ensemble-SVM & PLS-SVM OA= 88.99\%, Ensemble-SVM OA= 92.51\% \\

Improving Green Pepper Segmentation & \cite{liu2023pixelwise} & CVNN based on 1D Fast Fourier Transform & Accuracy= 94.89\%, F1 Score= 89.55\% \\

\bottomrule
\end{tabular}
\end{center}
\end{table}
\subsection{Contaminants and nutrient estimation}

Estimating nutrients and biomass in crops assists in the classification of crop conditions and various soil-characterized crop categories to promote agricultural development for farmers or others. Rapid determination of the nutritional content of lettuce cultivars grown hydroponically.
HSI technology can assist in the estimation of contaminants and nutrients in crops by collecting detailed spectral data that can be used to identify specific chemicals and substances in the crops and diagnose potential nutrient deficiencies \cite{watt2019application} and can also detect contaminants, such as heavy metals, that may be present in the crops. Additionally, HSI technology can be used to detect trends in nutrient uptake and availability \cite{haboudane2008remote}, which can be valuable for nutrient management and fertilization practices. The technology can also be used to monitor the impact of contaminants on crop health and estimate potential yield losses, providing a useful tool for precision agriculture and ensuring the safety and quality of crops for human and animal consumption. Table \ref{contnutr} summarizes some of the most important research on HSI for Contaminants and nutrient estimation applications.

\begin{table}
\begin{center}

\caption{An overview of the most significant research on the use of HSI for estimating contaminants and nutrients.}\label{contnutr}

 \begin{tabular}{p{1.5in}p{0.5in}p{1.4in}p{2.2in}}
\toprule%
Objective & Article & Method & Results \\ \midrule

Estimation of wheat nitrogen content & \cite{al2019spectral} & PLS-R & average error =0.83\%DM \\

Assessment of chlorophyll content & \cite{gao2021improvement} &  ROI-ALR- RF -PLSR with ROI-OMR & Rv2 = 0.52 and RMSE=3.61 \\

Estimation of the starch content in an individual kernel & \cite{liu2020determination} &  ANN, PLSR,  adaptive reweighted sampling & Rp2= 0.96 RMSEP= 0.98 \\

Estimation of nitrogen content in leaves of olive trees & \cite{rubio2021predicting} & PLSR, SVM, LNC, SM & R2=0.72, R2cv=0.71, R2=0.64, Rp2=0.63 \\

Various minerals types classification & \cite{agrawal2023deep} & mineral-CNN-LSTM based on 1D-CNN and LSTM and mineral-ResNet based on 1D-CNN, LSTM & mineral-ResNet OA= 92.16\% and kappa value of 0.89, mineral-CNN-LSTM OA= 91.71\% and kappa value of 0.88 \\

\bottomrule
\end{tabular}
\end{center}
\end{table}

\subsection{Plant disease monitoring and invasive plant species}

The appearance of several pests and illnesses in the crops poses serious challenges to farmers. Some of the frequent causes of illness infections include nematodes, bacteria, viruses, and fungi. Due to unawareness of crop diseases and the need for expert support and advice, farmers have historically avoided diagnosing or suspecting the majority of infections. In order to prevent actual damage to the crops, disease infections should be detected early on. HSI technology can assist in plant disease monitoring and invasive plant species detection by analyzing the reflectance spectra of the plants in multiple narrow, contiguous wavelength bands. This technology can detect subtle changes in the chemical and physical properties of plants that are not visible to the human eye \cite{moghadam2017plant}, such as changes in chlorophyll content or leaf water content. This information can then be used to identify plant stress and disease symptoms \cite{mahlein2012hyperspectral}, such as discoloration or wilting, or to distinguish between invasive plant species and native species based on differences in their spectral signatures. This allows for early detection and monitoring of plant diseases and invasive species, which can lead to improved management strategies and outcomes. Table \ref{plantdes} summarizes some of the most important research on HSI for Plant disease monitoring and invasive plant species applications.

\begin{table}
\begin{center}

\caption{overview of the most significant research conducted in the field of agricultural applications of HSI for the monitoring of plant diseases and invasive plant species}\label{plantdes}

 \begin{tabular}{p{1.5in}p{0.5in}p{1.4in}p{2.2in}}
\toprule%
Objective & Article & Method & Results \\ \midrule

Detection of target spot and bacterial spot diseases in tomato & \cite{abdulridha2020detection} & MLP Neural Network, Stepwise Discriminant Analysis (STDA) & Classification accuracy =99\% for both target spot (TS) \\

classification of the asymptomatic biotrophic phase of PLB disease & \cite{qi2023field} & 2DCNN and 3DCNN with attention networks & accuracy= 79.0\% F1 score= 0.83 \\

3D deep learning for plant disease recognition & \cite{nagasubramanian2019plant} & DCNN & Classification accuracy= 95.73\%, F1 score = 0.87 \\

Identification of red-berried wine grape infected with grapevine leaf-roll disease & \cite{gao2020early} &  Monte-Carlo, SVM, GLD & classification accuracy=89.93\% \\

aflatoxin B1 detection & \cite{zhu2023quantitative} & Dual-branch CNN & classification accuracy 91.30\% \\

assessment of weed competitiveness in maize farmland ecosystems & \cite{lou2022hyperspectral} & 3D-CNN & RMSE = 0.106 and 0.152 using 13 feature bands \\

Disease detection of basal stem rot & \cite{yong2023automatic} & Mask RCNN and VGG16 & classification accuracy 91.93\% \\

\bottomrule
\end{tabular}
\end{center}
\end{table}

\subsection{Plastic pollution}
Even if not strictly correlated with agriculture, plastic pollution has become one of the most emergent issues threatening aquatic and terrestrial ecosystems; so, its implications in agriculture are intrinsic.

The detection of plastic in the wild is a challenging area. As evident, the size of the monitored plastics is related to the resolution of the sensor, i.e. the flight altitude of the UAV. Microplastic can be detected only in the laboratory, with a distance between the target and the sensor in the range of centimeters \cite{zhu2021optimization}. On the other hand, macroplastics can be detected at a higher distance; basically, many approaches make use of satellite images \cite{themistocleous2020investigating}, \cite{topouzelis2020remote}. These approaches suffer from the main limitations of this kind of sensor: low spatial resolution, fixed time samples, rigid protocols for data access, and no customizable acquisition campaigns. On the other hand, they can count on high-quality data in terms of spectral resolution. The target of such approaches is large plastic waste detection, and it is performed by means of some specific spectral indices such as Reversed Normalized Difference Vegetation Index (RNDVI), Normalized Difference Water Index 2 (NDWI2), Plastic Index (PI), and Floating Debris Index (FDI), as reported in \cite{page2020identification} and \cite{biermann2020finding}.

The possibility to mount a hyperspectral sensor on a UAV opens a large field of applications oriented to small plastics detection \cite{cortesi2022uav}. In this scenario, the limitations of satellite data are overcome, and the increase in the spatial resolution gives a comfortable contribution to developing on-demand models able to perform path planning, data acquisition, and data processing in a compact time lapse. These advantages are indisputable and contribute to the growing use of these methodologies. In \cite{gonccalves2022operational} and \cite{balsi2021high} authors propose an approach to detect litter in a marine environment by using limited samples of the spectral bandwidths. The processing of acquired data can be performed by random forest classifiers \cite{cortesi2022uav} or, recently, by deep learning-based approaches \cite{maharjan2022detection}. The use of information coming from a sensor working in Short-Wave InfraRed (SWIR) bandwidth is proposed in \cite{cocking2022aerial}. This area of the spectrum is mainly used to distinguish between different kinds of plastics (i.,e PET, PVC, and so on), while the detection can be performed also in different bandwidths, specifically the range 600-900nm is suitable for this.
In the last years, the use of plastic in agriculture has become massive, mainly due to an increase in plastic-covered greenhouse farming areas and plastic-mulched farmlands; this makes it necessary to develop tools and approaches for the automatic detection and monitoring of such areas, for sustainable development of horticulture, including high-quality agricultural production, and reduced pollution \cite{veettil2023remote}, \cite{Levin2007}, \cite{Roldan2015}.

\section{Discussion}

HSI technology offers considerable advantages over conventional nondestructive testing methods. The inaccuracy caused by manual processes, instrument usage, and various reagent preparation steps that are included in classical detection methods is unpredictable. The hyperspectral data of the sample must first be extracted using HSI technology by taking pictures of the sample during detection. This data can then be merged with a machine-learning algorithm to determine the sample quality. In order to deliver accurate, quick, and nondestructive detection, hyperspectral detection eliminates errors caused by external variables including reagents, instruments, and operators. The early utilization of hyperspectral data encountered limitations. Initially, researchers followed a conventional approach of pre-processing (if necessary), extracting, and selecting discriminative features before applying a classifier to identify land cover groups. Feature extraction techniques such as PCA, ICA, and wavelets were emphasized, but these classic mathematical methods were insufficient for handling the massive amount of data involved in HSI classification. They were not capable of accurately predicting multiclass problems and had difficulty with feature selection and storage. Consequently, researchers faced challenges in analyzing, processing, and classifying HSIs. However, the development of ML/DL technologies has provided researchers with new avenues for addressing these challenges. Despite this progress, analyzing and extracting information from HSIs remains challenging due to the large number of highly correlated bands and high spatial-spectral features embedded in the electromagnetic spectrum. Finding appropriate technologies for classifying these interconnected, high-dimensional images requires sufficient quality-labeled data, and unsupervised approaches often result in inaccurate results due to a lack of coherence between spectral clusters and target regions.

\section{Conclusion}
This review paper comprehensively presents the individual information for each method, including their performance, research gaps, and achievements. Additionally, it introduces a novel research methodology that distinguishes this work from others. Through an in-depth examination of each methodology, significant inferences are drawn, enhancing the novelty of the work. Furthermore, the paper highlights the applications of HSI technology in agriculture and details the current research scenario on HSI classification. The paper also covers some of the recently developed techniques that may be particularly useful in future research.


\subsection{Limitations and Future Scope}

This article presents an overview of the technologies and procedures used for HSI classification from its inception to the present day. Despite the significant challenges associated with processing high-band data, researchers have made significant progress in this field over the last decade, improving existing techniques and developing new ones. With advancements in technology and the introduction of machine learning, HSI classification has become more accurate than traditional and contemporary state-of-the-art methodologies. As a result, deep learning has emerged as the most prominent tool for HSI classification in the past decade, leading to a greater exploration of remote sensing and space imagery features.

HSI exploration has opened up new avenues for research with numerous real-world applications. While HSI spectral bands offer certain advantages over multispectral and RGB imaging, they also come with limitations and disadvantages. Therefore, it is crucial to address the challenges associated with HSI analysis, some of which are listed below:

\begin{itemize}
    \item Data complexity and high cost.
    \item Despite the availability of high spectral information, the low spatial resolution of HSI data presents irregularities and challenges in interpretation.
    \item Publicly available datasets often have limited training samples, posing a challenge for HSI analysis.
    \item Target detection remains a significant difficulty in HSI due to the unpredictable nature of target and background spectra, making it challenging to develop efficient target detection algorithms.
\end{itemize}

This paper provides an in-depth analysis of popular deep learning methods used for HSI classification and evaluates their effectiveness. The techniques discussed below offer researchers new and improved approaches to HSI analysis:


\begin{itemize}
    \item Meta-learning, which involves creating algorithms that use the same datasets and combine predictions from various models, is an unexplored area of research in HSI classification.

    \item Newer datasets, such as the RIT-18 Remote Sensory Dataset \cite{kemker2018algorithms} and the olive \cite{dominguez2023field} and grape berry \cite{ryckewaert2023dataset} datasets , need to be used to test existing studies and improved techniques. Although the RIT-18 dataset is a multispectral dataset, this offers reliable and more versatile classification methods.
    \item Automatic parameter selection and optimization for DL and other approaches can benefit from an effective evolutionary technique or genetic algorithm.
    \item Data cubes offer a wealth of information about a scene, but the strong correlation between bands can result in duplicate information. Using automatic protocols to diagnose redundant information can improve classification accuracy.
    \item Creating cheap, compact, and lightweight mobile HSI setups, such as hardware like filters, sensors, and lighting sources, is an exciting research topic.
    \item Recent approaches to pattern recognition can be explored to learn information from data cubes more effectively and efficiently, as spectral signatures are obtained by extracting and analyzing data from high-dimensional data cubes.
\end{itemize}

\section*{Acknowledgement}
This work was supported by the Ministry of Enterprises and Made in Italy with the grant ENDOR "ENabling technologies for Defence and mOnitoring of the foRests" - PON 2014-2020 FESR - CUP B82C21001750005.


\end{document}